\documentclass[journal]{IEEEtran}

\ifCLASSINFOpdf
\else
   \usepackage[dvips]{graphicx}
\fi
\usepackage{url}

\usepackage{graphicx}

\usepackage{subfigure}  

\usepackage{color}
\usepackage{multirow}
\usepackage{amsmath}
\usepackage{booktabs}
\usepackage{float}
\usepackage{bm}
\usepackage{wrapfig}
\usepackage{mathrsfs}
\usepackage{bbding}
\usepackage{amssymb}
\usepackage{caption}

\usepackage{amsfonts}       
\usepackage{nicefrac}       
\usepackage{microtype}      
\usepackage{xcolor}

\usepackage{array} 
\usepackage{pifont}

\usepackage{bbding}

\usepackage{colortbl}
\definecolor{mygray}{gray}{.9}
\definecolor{mypink}{rgb}{.99,.91,.95}
\definecolor{mycyan}{cmyk}{.3,0,0,0}

\begin{document}

\title{COMNet: Co-Occurrent Matching for Weakly Supervised Semantic Segmentation}

\author{Yukun Su, Jingliang Deng, Zonghan Li

\thanks{Yu. S, Jing. D and Zong. L are with the School of Software Engineering, South China University of Technology.}
}

\markboth{Journal of \LaTeX\ Class Files, Vol. 14, No. 8, April 2022}
{Shell \MakeLowercase{\textit{et al.}}: Bare Demo of IEEEtran.cls for IEEE Journals}
\maketitle

\begin{abstract}
Image-level weakly supervised semantic segmentation is a challenging task that has been deeply studied in recent years. Most of the common solutions exploit class activation map (CAM) to locate object regions. However, such response maps generated by the classification network usually focus on discriminative object parts. In this paper, we propose a novel $\textbf{C}$o-$\textbf{O}$ccurrent $\textbf{M}$atching $\textbf{N}$et (COMNet), which can promote the quality of the CAMs and enforce the network to pay attention to the entire parts of objects. Specifically, we perform inter-matching on paired images that contain common classes to enhance the corresponded areas, and construct intra-matching on a single image to propagate the semantic features across the object regions.
The experiments on the Pascal VOC 2012 and MS-COCO dataset show that our network can effectively boost the performance of the baseline model and achieve new state-of-the-art performance.
\end{abstract}

\begin{IEEEkeywords}
Weakly Supervised, Co-Occurrent, Matching, Segmentation.
\end{IEEEkeywords}

\IEEEpeerreviewmaketitle

\section{Introduction}

\IEEEPARstart{S}{emantic} segmentation is a foundation in the computer vision field, which aims to predict the pixel-wise classification of the images and it enjoys a wide range of applications, such as image analysis~\cite{chen2017deeplab, su2022epnet}, geological inspection~\cite{dai2021transformer}, and auto vehicle~\cite{li2021bifnet, su2023occlusion}, etc. Recently, benefiting from the deep neural networks,  modern semantic segmentation models~\cite{chen2018encoder,lin2016efficient,su2023unified} have achieved remarkable progress with massive human-annotated labelled data. 
However, collecting pixel-level labels is very time-consuming and labour-intensive, which shifts much research attention to weakly supervised semantic segmentation (WSSS).

There exist various types of weak supervision for semantic segmentation like using bounding boxes~\cite{dai2015boxsup,khoreva2017simple}, scribbles~\cite{lin2016scribblesup,vernaza2017learning}, points~\cite{bearman2016s}, and image-level labels~\cite{ahn2018learning,ahn2019weakly,wang2020self,zhang2020splitting,su2021context,huo2022dual,cao2023semantic}. Among them, image-level class labels have been widely used since they demand the least annotation efforts and are already provided in existing large-scale image datasets.
To tackle the task in WSSS, the mainstream pipelines are based on visualization network CAM~\cite{zhou2016learning}, which first trains a classifier network with image-level labels and discover discriminative regions that are activated for classification. Second, by expanding the seed areas using different techniques~\cite{kolesnikov2016seed,ahn2018learning,ahn2019weakly}, we can obtain the pseudo-masks as the ground-truth for training a standard supervised semantic segmentation model.

In this paper, we focus on weakly supervised segmentation with only image-level labels are available during the training process. However, CAMs tend to focus only on the most discriminative regions with strong semantic cues instead of the whole object regions. which will hurt the subsequent steps and yield low-quality pseudo-masks. 
To this end, we proposed a novel \textbf{C}o-\textbf{o}ccurrent \textbf{M}atching \textbf{N}et termed as \textbf{COMNet} in this paper to tackle the drawbacks in conventional CAMs. 
Specifically, instead of using the single training images independently, we adopt paired images that contain common classes for training. We first encode the paired images with the share-weight network (i.e., ResNet50~\cite{resnet}) into feature representations, and then we perform a cluster inter-matching for these paired features in an unsupervised manner to yield the optimal masks for paired images, which are then used to enhance the corresponded representations.

To better improve the response maps, we further propose an intra-matching strategy within the single images. Based on the observation that the semantic information (i.e., color and texture) of the whole regions of the objects should be highly relevant. Therefore, we can propagate the semantic information from the strong-activated regions to the weak-activated regions by utilizing pixel's $K$ neighbourhood in feature spaces.

By combining these two proposed matching modules, we train our \textbf{COMNet} in an end-to-end manner. Experimental results show that our method outperforms the baseline methods and achieves new state-of-the-art performance by 67.1\% mIoU on the val-set and 67.6\% mIoU on the test-set of PASCAL VOC 2012~\cite{voc}.
The main contributions of our work are the following:
\begin{itemize}
	\item We propose an end-to-end training COMNet for weakly supervised segmentation tasks from a novel matching perspective.
	\item The proposed inter-matching module can help the network enhance the co-occurrent regions and to discover common object areas. To further improve object regions predictions, the intra-matching module is proposed to propagate semantic information in feature spaces to expand object activated regions.
	\item Experimental results on the PASCAL VOC 2012 and MS-COCO datasets show that our method outperforms the baseline method and achieves new state-of-the-art performance.
\end{itemize}

\begin{figure*}
	\begin{center}
		\centering
		\includegraphics[width=6.3in]{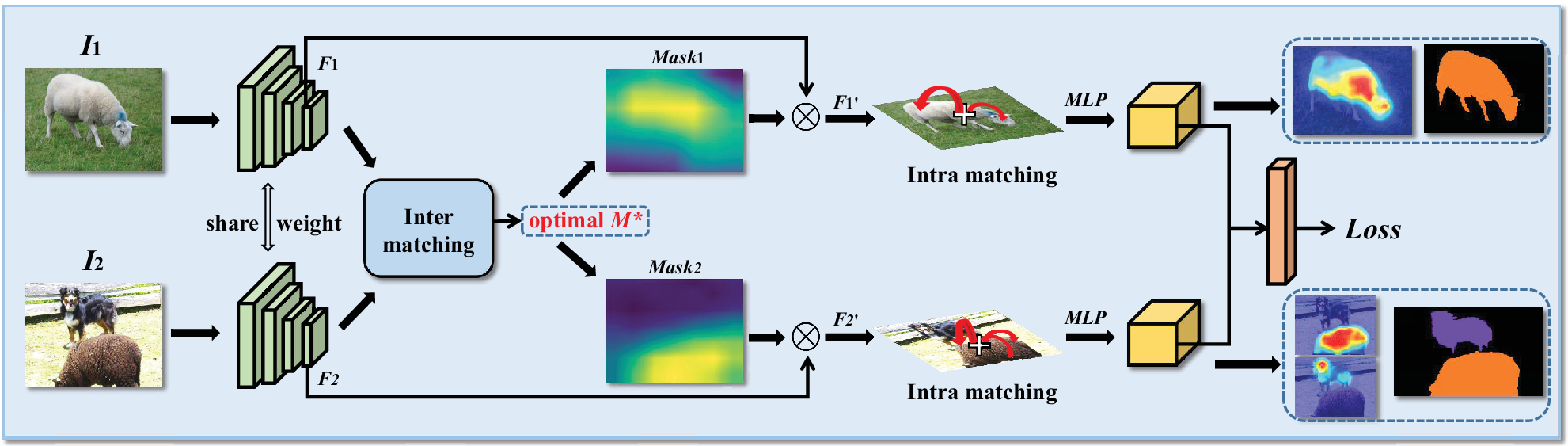}
	\end{center}
	\caption{\textbf{Overview of our method}. Paired training images that contain common classes are first encoded into feature representations by a share-weight network. The optimal solution masks from the inter-matching module are then used to reinforce the co-occurrent object regions. Then the intra-matching module further enhances the learning features by propagating semantic information across regions. The final standard multi-label classification loss $\mathcal{L}_{cls}$ is used to update the gradient propagation to train the network in an end-to-end manner.}
	\label{fig2}
\end{figure*} 

\section{Related Work}
\label{sec:guidelines}

\subsection{Weakly supervised semantic segmentation}

Image labels as the weak supervision for segmentation have been widely studied in the past few years. Many approaches~\cite{affinitynet,irn} use CAM~\cite{cam} to mine the object seed regions by predicting image labels.  
For example, \cite{yu2015multi} enlarges object regions by fusing the different discriminative regions generated by convolutional layers with different dilation rates. Different convolutional layers are expected to capture different discriminative object parts. \cite{wei2017object} expands the seed regions generated with CAMs by erasing the discriminative regions detected by the region-mining model and then re-train the model with the erased images. Researchers also propose to expand the discriminative regions by seeking external techniques. \cite{hou2018self} incorporates saliency maps into the network training to better inference the background and foreground object regions. \cite{ahn2018learning} utilizes the inter-pixel relations to refine the boundaries of the object areas. There are also works~(\cite{jin2017webly,shen2018bootstrapping}) utilizing external web images to improve the segmentation accuracy. Other methods adopting motion videos~(\cite{hong2017weakly}), instance saliency mask~(\cite{fan2018associating}) also yield promising results.

\subsection{Co-occurrent Matching}
However, seeking information based on a single image is limited in capturing substantial semantic context. Therefore, there are also some researches endeavour to aggregate self-attention modules in the WSSS framework. \cite{fan2020cian} proposes a cross-image attention module to learn activation maps from two different images containing the same class objects with the guidance of saliency maps. More recent approaches follow the self-supervised paradigm to acquire additional supervision~(\cite{shimoda2019self,wang2020self}),
or rely on Siamese networks to capture semantic relations between a pair of images~(\cite{sun2020mining}).
Although these works leverage the multi-images semantic context for WSSS, they fail to model the feature-level relationship explicitly and they cannot reasonably explain how to enhance complementary information across-images. Thus, exploring the matching methodology to enhance object region mining is a new trend and it can also meet the interpretability of artificial intelligence.
Matching has been widely used and studied in computer vision tasks, which explores the shape correspondences within a pair of images. \cite{su2020human} uses kernel graph matching to solve the video classification task. \cite{zhang2020deepemd} proposes to use the Earth Mover’s Distance to find the matching regions between two images for few-shot classification. Graph matching also has been used in image segmentation. \cite{kainmueller2014active} and \cite{martins2011segmentation} segment the similar images with the graph matching. \cite{zhang2019pyramid} uses the bipartite graph with attention mechanism to establish cross-image region correspondence for few-shot image segmentation. ~\cite{su2022self} explores cluster matching for object detection.
By exploring image matching for weakly supervised semantic segmentation, we can further discover more object regions of the networks and also can extend pair-wise matching to group-wise matching to mine richer semantic information.

\section{Method}
This section details our \textbf{COMNet} structure. The overview of our method for weakly supervised semantic segmentation is shown in Fig~\ref{fig2}. Our proposed method aims to generate high-quality CAMs. Based on the response maps generated by our method, we adopt the random walk method~\cite{irn} to refine the maps as pixel-wise pseudo ground truths for semantic segmentation. 
Next, we will give more details of how we explore paired images inter-matching and single images intra-matching to improve the response maps and generate final semantic segmentation results.

\subsection{Paired Images Inter-Matching}

\begin{figure}
	\begin{center}
		\centering
		\includegraphics[width=3.3in]{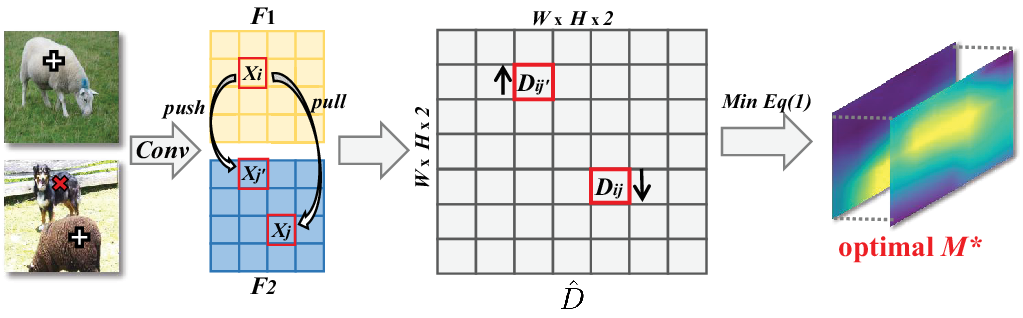}
	\end{center}
	\caption{\textbf{Illustration of the paired images inter-matching module}. The objective is to pull the features of the same classes together while pushing away the features of different classes. The paired images are encoded into feature representations as $F_1$ and $F_2$. Then they are reformulated into an affinity matrix $\hat{D}$, where the entry of the matrix represents the distance among the features. By minimizing the Eq (\ref{eq1}), we can obtain the optimal solution mask ${M*}$.}
	\label{fig3}
\end{figure}

Given a paired images $\mathcal{I}$ = $\left\{ I_1, I_2 \right\}$ that contain co-objects of specific categories, we first encode two images into feature representations as $\left\{ F_1, F_2 \right\}$ $\in$ $\mathbb R^{w \times h \times c}$  with a share-weight convolution neural network. Our objective is to learn the co-occurrent object masks and then reinforce the representations to highlight the object regions. However, since we do not have extra supervision for training, obtaining co-occurrent masks by using a supervised learning method is unfeasible. Therefore, we propose an unsupervised cluster matching approach, which aims to minimize the distances between the same classes features while maximizing the distances between different classes features. Fig~\ref{fig3} shows the diagram of the paired images inter-matching module.

Concretely, we use $X = [x_1, ... , x_{2wh}]$ to represent the paired images features, this can be mathematically done by concatenating $\left\{ F_1, F_2 \right\}$ and then reshape them. Here, if $x_i$ belongs to the corresponding classes, we assign it to foreground ($fg$) features. Otherwise, we assign it to background ($bg$) features.
To this end, similar to ~\cite{su2022self}, the cluster inter-matching objective can be defined as follows:

\begin{equation}
\mathcal{L}_{mask} = \sum_{i,j \in fg}^{}D_{ij} + \sum_{i,j \in bg}^{}D_{ij} - \sum_{i \in fg, j \in bg}^{}D_{ij} - \sum_{i \in bg, j \in fg}^{}D_{ij} ,\tag{1}
\label{eq1}
\end{equation}

here we use ${\ell_2}$- distance to represent $D_{ij} = ||x_i - x_j||^2_2$ between feature $i$ and feature $j$.

Note that solving Eq (\ref{eq1}) is a discrete problem, so we introduce a cluster vector to reformulate the function. The optimization learning problem is then transformed into an integer programming problem. Its continuous relaxation has a closed-form solution. The learning parameter indicates whether the image pixel belongs to the corresponding classes.

Let the cluster indicator vector $M = [m_1, ..., m_{2wh}]^\mathit{T}$ be: 

\begin{equation}
\begin{split}
M_{(i)}= \begin{cases}
 1/\sqrt{2wh}  &, \ \ if \ \  i \in fg,  \\
-1/\sqrt{2wh}  &, \ \ if \ \  i \in bg. 
\end{cases} \end{split}  \tag{2}
\label{eq2}
\end{equation}

The indicator vector satisfies the sum-to-zero and normalization conditions: $\sum_{}^{}M_{(i)} = 0$, $\sum_{}^{}{M_{(i)}}^2 = 1$. The objective function in Eq (\ref{eq1}) can be rewrite in matrix form as:

\begin{equation}
\mathcal{L}_{mask} = 2whM^\mathit{T}DM.\tag{3}
\label{eq3}
\end{equation}

Note that we normalize the channel features $\mathit{s.t.} ||x_i||^2_2 = 1$, therefore, $D_{ij} = 2-2x_i^\mathit{T}x_j$. To this end, by removing the trivial constant $2wh$, $\mathcal{L}_{mask}$ can be reformulated as:

\begin{equation}
\mathcal{L}_{mask} = -M^\mathit{T}\hat{D}M,\tag{4}
\label{eq4}
\end{equation}
where $\hat{D}= X^\mathit{T}X-I$ and $I$ denotes an all-ones matrix, relaxing the elements in $M$ from binary indicator values to continuous values in [-1,1] subject to $||M||^2_2$ = 1. The optimal solution $M* =$ argmin $(\mathcal{L}_{mask}) $ satisfies the theorem in ~\cite{ding2004k}.

Finally, the optimal co-occurrent mask $M* \in \mathbb R^{2wh}$ is reshaped into $\left\{ M_1, M_2 \right\}$ $\in$ $\mathbb R^{w\times h}$. We then re-weight all the feature representations in the matching positions, which can be done by multiplying the original feature representations with the masks, and the new feature representations can be defined as follows:

\begin{equation}
\begin{split}
F_{1(i)}^{'}= \begin{cases}
F_{1(i)} \times \alpha  &, \ \ if \ \  M_{1(i)} > 0,  \\
F_{1(i)}\times 1  &, \ \ if \ \  M_{1(i)} < 0,
\end{cases} \end{split}  \tag{5}
\label{eq5}
\end{equation}
where $\alpha$ is a hyper-parameter, and its value is set to $>$ 1. For simplicity, $F_{2}^{'}$ is operated in the same way like $F_{1}^{'}$.

\vspace{1ex}

\textbf{Derivation of the function.} We present the detailed explanation of the $L_{mask}$ in the followings.

{\tiny{
$$
		D=	\left|
				\begin{array}{ccccc}
					d_{11} &\cdots  &\cdots& \cdots  & d_{1,2wh}\\
					\vdots & \ddots &      &  		 &\vdots \\
					\vdots &  		& d_{i,j} & 	 &\vdots \\
					\vdots & 		&   &\ddots  	 &\vdots \\
				    d_{2wh,1} &\cdots  & \cdots &\cdots & d_{2wh,2wh} 
				\end{array}
			\right|
		M=\left|
				\begin{array}{c}
					m_{1}\\
					\vdots\\
					m_{i}\\
					\vdots\\
					m_{2wh}
				\end{array}
		  \right|
	$$

	\begin{align}
		L_{mask}
		& = 2whM^TDM \\ 
		& = 2whM^T\left|
						\begin{array}{ccccc}
							d_{11} &\cdots  &\cdots& \cdots  & d_{1,2wh}\\
							\vdots & \ddots &      &  		 &\vdots \\
							\vdots &  		& d_{i,j} & 	 &\vdots \\
							\vdots & 		&   &\ddots  	 &\vdots \\
							d_{2wh,1} &\cdots  & \cdots &\cdots & d_{2wh,2wh} 
						\end{array}
					\right|	M	\\
		& = 2wh\left|
					\begin{array}{ccccc}
						\sum\limits_1^{2wh} M^T_{1,i}d_{i,1}& \cdots & \sum\limits_1^{2wh} M^T_{1,i}d_{i,j}& \cdots &\sum\limits_1^{2wh} M^T_{1,i}d_{i,2wh} 
					\end{array}
			  \right| M	\\ 
		& = 2wh(M_{1,1}\sum\limits_1^{2wh} M^T_{1,i}d_{i,1}+\cdots+M_{j,1}\sum\limits_1^{2wh} M^T_{1,i}d_{i,j}+\cdots \\ 
            & +M_{2wh,1}\sum\limits_1^{2wh} M^T_{1,i}d_{i,2wh})	\\ 
		& = 2wh \sum\limits_1^{2wh}M_{j,1}\sum\limits_1^{2wh} M^T_{1,i}d_{i,j}	\\ 
		& = \sqrt{2wh}\sum\limits_1^{2wh}M_{j,1}(\sum_{i\in fg}d_{i,j}-\sum_{i\in bg}d_{i,j})	\\
		& = \sum_{i,j\in fg} d_{i,j}+\sum_{i,j\in bg} d_{i,j}-\sum_{i\in fg,j \in bg} d_{i,j}-\sum_{i\in bg,j \in fg} d_{i,j}
	\end{align}
	
	\begin{align}
		\hat{D}
		& = \left|
				\begin{array}{ccccc}
					x_1^Tx_1 &\cdots  &\cdots& \cdots  & x_1^Tx_{2wh}\\
					\vdots & \ddots &      &  		 &\vdots \\
					\vdots &  		& x_i^Tx_i & 	 &\vdots \\
					\vdots & 		&     &\ddots  	 &\vdots \\
					x_{2wh}^Tx_1 &\cdots  & \cdots &\cdots & x_{2wh}^Tx_{2wh} 
				\end{array}
			\right|-I \\
		& = \left|
				\begin{array}{ccccc}
					x_1^Tx_1-1 &\cdots  &\cdots& \cdots  & x_1^Tx_{2wh}-1\\
					\vdots & \ddots &      &  		 &\vdots \\
					\vdots &  		& x_i^Tx_i-1 & 	 &\vdots \\
					\vdots & 		&     &\ddots  	 &\vdots \\
					x_{2wh}^Tx_1-1&\cdots  & \cdots &\cdots & x_{2wh}^Tx_{2wh}-1 
				\end{array}
			\right| \\
		& = -\frac{1}{2}D
	\end{align}
 }}

\subsection{Single Images Intra-Matching}\label{intra}

To further discover the object regions, we adopt intra-matching proposed in~\cite{su2023unified} within the single images to propagate the feature representations. Traditional matching usually only considers its nearest points (e.g., the point neighbourhood refers to the pixels that are horizontally and vertically connected), however, we argue that this will make the networks to focus on local areas.  
Contrastingly, based on the observation that the semantic information of the whole objects shares the feature similarity and consistency. We drive the networks to automatically match the points from the entire feature maps to better expand the activated response maps since the weak-activated region points can be matched with some high-activated region points. With this in mind, Fig~\ref{fig4} shows the diagram of the single images intra-matching module and we have the following steps:

\begin{figure}
	\begin{center}
		\centering
		\includegraphics[width=3.3in]{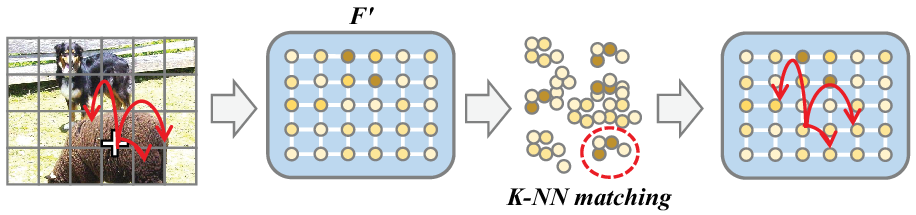}
		\includegraphics[width=3.3in]{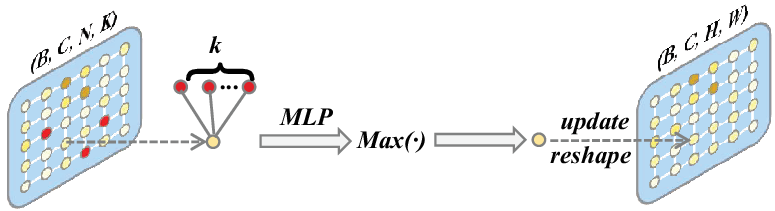}
	\end{center}
	\caption{Illustration of the single images intra-matching module. \textbf{Top}: The points indicate the pixel features of the images. For each point, it will match with its top-$k$ potentially corresponding points. \textbf{Bottom}: The process of updating each point after it is matched to its corresponding points.}
	\label{fig4}
\end{figure}

\subsubsection{Top-k Matching}
Given a single image feature $F^{'} \in {\mathbb R^{w\times h \times c}}$ after adopting inter-matching, we first reshape it into $c \times N$ dimension, where $N = w\times h$ denotes the number of feature points.
In order to calculate the pair-wise semantic similarity of all the feature points, we then establish a similarity matrix $G \in \mathbb R^{N\times N }$, where the $(i, j )$-th entry $G_{ij}$ represents the relation between point-$i$ and point-$j$. Here we use the ${\ell_2}$-distance as our metric, $G$ can be formulated as follows:

\begin{equation}
{G} = F^{'\mathit{T}}F^{'}.\tag{6}
\label{eq6}
\end{equation}

To avoid matching the points with themselves, the diagonal elements of the matrix are set to {-Inf}. Then we apply an arg top-$k$ operation to the matrix, and it will output a tensor in $N \times K$ shape, which indicates that each point matches with its top-$k$ potentially corresponding points.

 \subsubsection{Layer Updating}
The details of the points updating layer are illustrated in Fig~\ref{fig4} bottom. After the top-$k$ matching operation, the feature points $F^{'} \in \mathbb R^{c\times N}$ will be transformed into $F^{'} \in \mathbb R^{c\times N \times K}$. In this way, we can view the feature points as a point set. 
For each local point set  $\left\{ p_1, p_2, ...,p_k \right\} \in \mathbb R^{c}$ according to the centroid point’s $K$ neighbors, ~\cite{pointnet} gives the theoretical foundation that: 

\begin{equation}
f(p_1, p_2, ...,p_k) = \gamma({\underset{i=1,..., k}{MAX}} \left\{ h(p_i) \right\} ),\tag{7}
\label{eq7}
\end{equation}
where $f: \mathcal{X}\to \mathbb R$  can be any continuous set function. $\gamma$ and $h$ are usually multi-layer perceptron (MLP) networks. $MAX$ is the element-wise maximum operation. This guarantees that the combination of MLP and symmetric function can arbitrarily approximate any continuous set function.

Following the rule of the above universal continuous set function approximator, each newly updated feature point $p^{'}$ will learn and aggregate all its potential correspondences as:

\begin{equation}
p^{'}_{(i)} = {\underset{j=1,..., k, i \notin j}{MAX}} \left\{ MLP(p_{(j)}) \right\}.\tag{8}
\label{eq8}
\end{equation}

Finally, $F^{'} \in \mathbb R^{c\times N \times K}$ will be transformed back to $F^{'} \in \mathbb R^{c\times N}$ by $MAX$ symmetric function, we then reshape it into $F^{'} \in \mathbb R^{c\times w\times h}$. After global average pooling, we use the multi-label soft cross-entropy classification loss to train the network in an end-to-end manner:

\begin{equation}
\begin{split}
\mathcal{L}_{cls} &=-\sum_{i}^{N}(y_i  log(\frac{1}{1 + exp(-x_i)}) \\ \notag
&+ (1-y_i)  log(exp(\frac{-x_i}{1 + exp(-x_i)}))),\end{split}\tag{9}
\label{eq9}
\end{equation}
where $x$ represents the predicted probability of class $i$ and $y_i$ denotes the ground-truth label of the $i$-th class.

\subsection{More In-Depth Discussion}\label{Discussion}

The above inter-matching module operates on a pair of two images. It can be easily extended to formulate relationships among multiple paired images.
Given a group of $N$ images $\mathcal{I}$ = $\left\{ I_1,..., I_N \right\}$ that contain common classes, we just need to reformulate the image number in Eq (\ref{eq2},\ref{eq3}) from 2 to $N$. Note that, as the group images $N$ increases, the matrix $\hat{D}$ gets larger,  the computational complexity and the time for matching will increase. In addition, the experimental results show that increasing $N$will not significantly improve the network performance, but will cause harm to our model. We conjecture that it is hard for the network to match with too many images that contain complex content and background information, which leads to a poor matching effect and affects the co-occurrent regions. Therefore, we adopt $N=2$, namely we use paired images as input to train our proposed method.

\section{Experiments and Analysis}
In this section, we first present the ablative analysis of the response maps generated by our method. Second, we show the semantic segmentation performance against the state-of-the-art approaches. More results can be found in the supplementary material.

\subsection{Dataset and Metric}

We evaluate the proposed method on the PASCAL VOC 2012~\cite{voc} segmentation benchmark for a fair comparison to previous approaches.
The official dataset separation has 1464 images for training, 1449 for validation and 1456 for testing.
Following the common practice, we take additional annotations to build an augmented training set with 10582 images presented in~\cite{hariharan2011semantic}. MS-COCO~\cite{lin2014microsoft} is a more challenging benchmark with 80 semantic classes. Since more complex contextual relations exist among these categories, it is interesting to examine the performance of our model in this dataset. Following~\cite{wang2020weakly}, we
use the default train/val splits (80k images for training and 40k for validation) in the experiment.
We use the standard mean Intersection-over-Union (\textbf{mIoU}) as the evaluation metric for all experiments.

\begin{table}[]
	\begin{center}
		\scalebox{1.0}{
			\begin{tabular}{c|c|c}
				\toprule  
				\toprule  
				Group images $N$ \ \ & \ \ Inter-matching times (s)\ \ & \ \ mIoU(\%)\ \ \\
				\midrule  
				baseline & - & {48.3} \\
				\midrule  
				\rowcolor{mygray} N = 2& 0.13 & \textbf{49.5} \\
				N = 3& 0.56 & {49.0} \\
				N = 4& 1.49 & {48.2} \\
				N = 5& 2.74 & {48.3} \\
				\bottomrule 
		\end{tabular}}
	\end{center}\caption{Experiments of different group images on time costing and seed area performance. Simply increasing $N$ cannot bring significant improvement.}\label{table1}
\end{table}

\begin{table}[]
	\begin{center}
		\scalebox{1.0}{
			\begin{tabular}{cccc}
				\toprule  
				\toprule  
				Baseline  & Inter-matching & Intra-matching & mIoU(\%) \\
				\midrule  
				\ding{51} &  & & 48.3 \\
				\ding{51}&  \ding{51} & & 49.5 \\
				\ding{51}&  & \ding{51} & 49.6 \\
				\midrule  
				\rowcolor{mygray} \ding{51}&  \ding{51} & \ding{51} & \textbf{50.8} \\
				\bottomrule 
		\end{tabular}}
	\end{center}\caption{Experiments of the two matching modules. Each matching module can independently boost the baseline model. By combining the proposed inter- and intra-matching modules, it can achieve the best performance.}\label{table2}
\end{table}

\begin{table}[]
	\begin{center}
		\scalebox{1.0}{
			\begin{tabular}{c|c|c|c|c}
				\toprule
				$\alpha$ \ & \ \ $ 1.2$ \ \  & \ \ $1.5$ \ \ & \ \ $1.8$ \ \ & \ \ $2.0$ \ \ \\
				\midrule
				mIoU(\%) \ & \ 49.2 \ & \ \textbf{49.5} \ & \ 49.3 \ & \ 49.2 \ \\
				\bottomrule
		\end{tabular}}
	\end{center}\caption{Comparison with different $\alpha$ in inter-matching mask enhancement.}\label{table3}
\end{table}

\begin{table}[]
	\begin{center}
		\scalebox{1.0}{
			\begin{tabular}{c|c|c|c|c|c|c}
				\toprule
				$k$ \ & \ \ $ 6$ \ \  & \ \ $8$ \ \ & \ \ $10$ \ \ & \ \ $12$ \ \ & \ \ $16$ \ \ & \ \ $32$ \ \ \\
				\midrule
				mIoU(\%) \ & \ 49.3 \ & \ \textbf{49.6} \ & \ {49.2} \ & \ 49.0 \ & \ 49.4 \ & \ 49.2 \ \\
				\bottomrule
		\end{tabular}}
	\end{center}\caption{Comparison with different $k$ potential points in intra-matching module.}\label{table4}
\end{table}

\subsection{Implementation Details}\label{Implementation Details}

In this work, we implement the proposed framework with PyTorch and train on a single Titan-X GPU. ResNet-50~\cite{resnet} is used as the backbone network (pre-trained on ImageNet~\cite{imagenet}). The input image was cropped into a fixed size of 512 $\times$ 512 using zero padding if needed.
We use stochastic gradient descent (SGD) optimizer with initial learning rate of 0.1 and batch size of 16 for the model. The learning rate decreases using polynomial decay $lr_{init}(1-itr/max\_itr)^\rho$ with $\rho$ = 0.9 at every iteration. The same data augmentation strategy (i.e., horizontal flip, random cropping, and color jittering) as in ~\cite{irn,affinitynet} is used in the training process. For a fair comparison, we choose DeepLab~\cite{deeplabv2} to train with the pseudo labels to achieve final segmentation results.
The fully-connected CRF~\cite{crf} is used to refine CAM, pseudo-mask, and segmentation mask with the default parameters.

\subsection{Ablation Studies}

To explore the components of our proposed method, we conduct extensive analysis to demonstrate how they help to improve feature representations and discover more parts of the objects. Here, all the experimental results are based on the PASCAL VOC training set for evaluating CAM seed area. \textbf{Baseline} model indicates the conventional CAM.

\vspace{1ex}

\noindent \textbf{The effect of group images on inter-matching.} Since we have discussed in Sec.\ref{Discussion} that our proposed method can adopt multiple images for inter-matching. Here, we first explore the effect of group images on network performance. As is shown in Tab~\ref{table1}, when $N = 2$, the inter-matching module can help the network to achieve the best performance over the baseline model. As the group images increase, more matching time is consumed. Besides, simply increasing group images can not bring significant improvement to the network and sometimes it gets poorer performance ($N=4$) than the baseline model.

\vspace{1ex}

\noindent \textbf{Comparison with baseline.} Tab~\ref{table2} gives an ablation study of each module in our approach. It shows that each of our matching module successfully gains improvement compared to baseline. Specifically, inter-matching module achieves \textbf{49.5\%} and intra-matching achieves \textbf{49.6\%} mIoU on PASCAL VOC training set. Furthermore, by combing these two proposed method, we can improve the performance up to \textbf{50.8\%}, which demonstrate that our proposed method can discover more potential object regions.

\vspace{1ex}

\noindent \textbf{The effect on hyper-parameter.} In our inter-matching module, we set $\alpha$ to re-weight the feature representations in the co-occurrent regions. As shown in Tab~\ref{table3}, when $\alpha$ is set to 1.5 it can achieve the best performance. In addition, in our intra-matching module, we further explore the effect of the top-$k$ number of points on network performance. As can be seen in Tab~\ref{table4}, when $k=8$, the intra-matching module can work out the best. We conjecture that too many points for matching will bring redundant information while too few points fail to provide useful semantic information.

\begin{figure}
	\begin{center}
		\centering
		\includegraphics[width=3.3in]{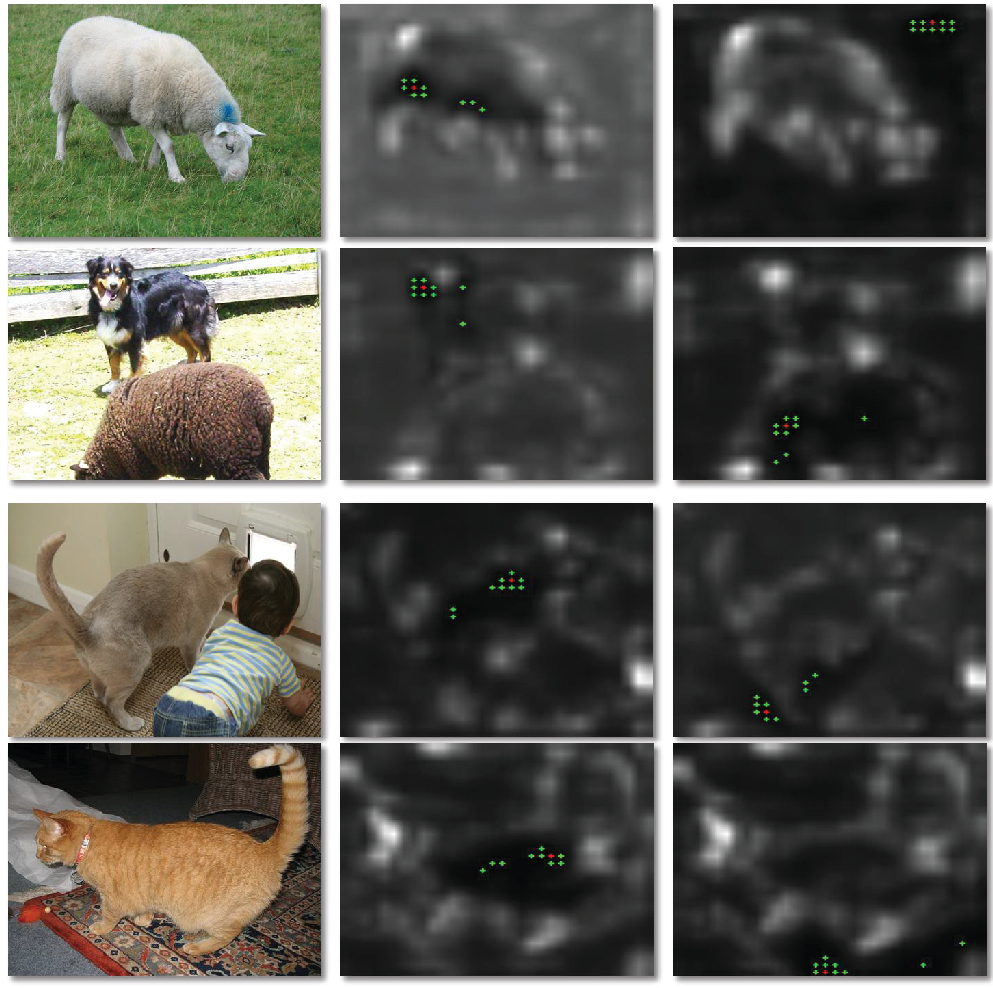}
	\end{center}
	\caption{The visualization of the single images intra-matching maps on foreground and background. The  {\color{red} red} cross denotes the selected pixel points, with similar feature representation in black color. The {\color{green} green} cross indicates its top-$k$ potentially corresponding points.}
	\label{fig5}
\end{figure}

\begin{figure}
	\begin{center}
		\centering
		\includegraphics[width=3.3in]{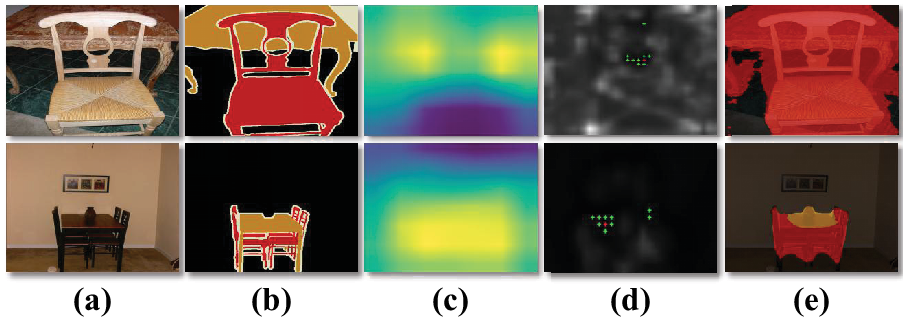}
	\end{center}
	\caption{Typical failure case. (a) Input images. (b) Ground truth. (c) Intermediate inter-matching masks. (d) Intra-matching point maps. (e) Our results.}
	\label{fig7}
\end{figure}

\begin{figure}
	\begin{center}
		\centering
		\includegraphics[width=3.3in]{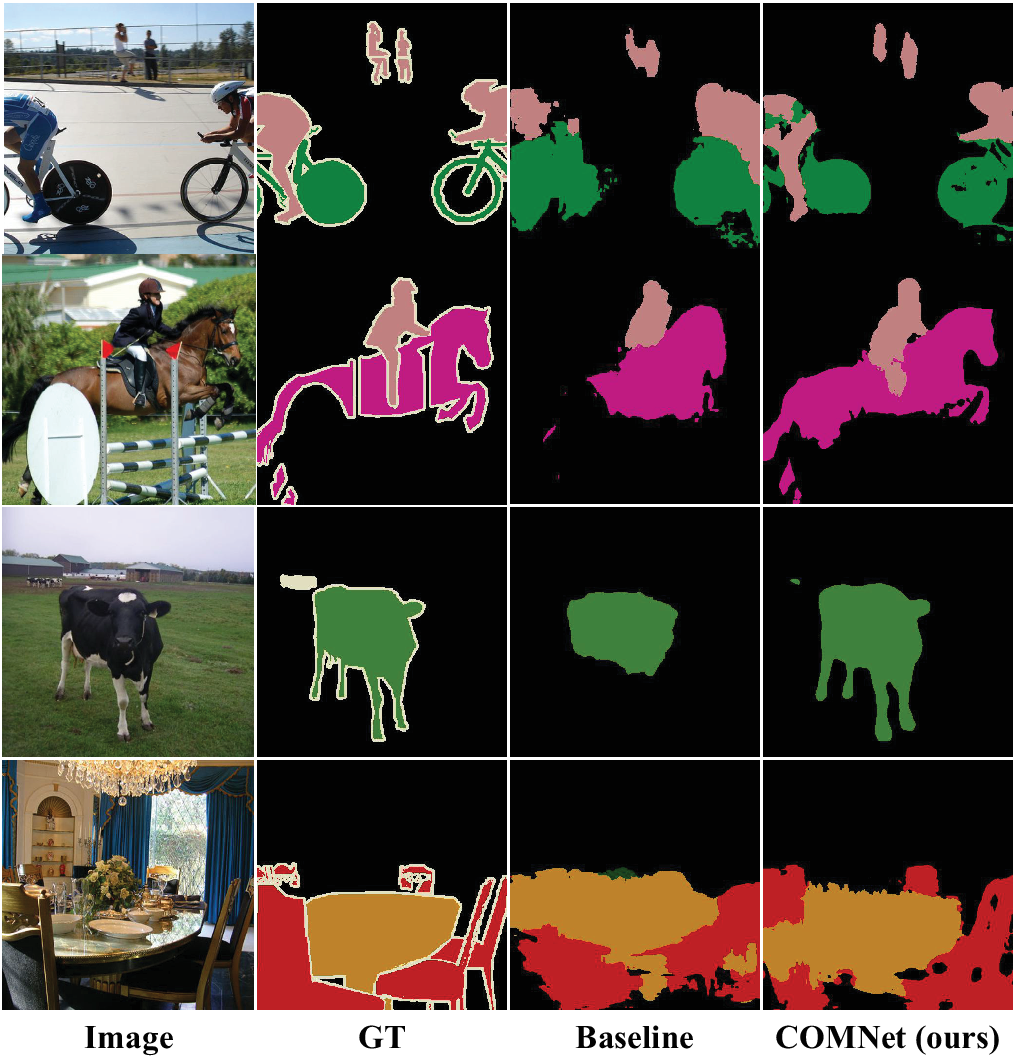}
	\end{center}
	\caption{Qualitative results on the PASCAL VOC 2012 validation set.}
	\label{fig8}
\end{figure}

\begin{table}[]
	\begin{center}
		\scalebox{1.0}{
			\begin{tabular}{ccc}
				\toprule  
				\toprule  
				Method  & CAM  & CAM + RW \\
				\midrule  
				IRN~\cite{irn}& 48.3  & 65.9 \\
				\midrule  
				\rowcolor{mygray}Ours&   50.8& 67.3 \\
				\bottomrule 
		\end{tabular}}
	\end{center}\caption{Performance comparison in mIoU (\%) for evaluating activation maps on the PASCAL VOC training sets.}\label{table6}
\end{table}

In order to have a more intuitive understanding of the intra-matching module, Fig~\ref{fig5} shows us the maps that reveal the distribution of the matching points. As can be seen, the selected points will diverge to find the matching points that are highly related to their semantic information. This can help the network to propagate the semantic information across the entire objects and to expand the response maps.

\subsection{Analysis of Pseudo Labels}

In Tab~\ref{table6}, we show the mIoU of the segments computed using the CAM on the training sets. We present the results after applying the refinement step to the activation map, i.e., CAM + random walk (CAM + RW). Tab~\ref{table6} shows that the synthesized segmentation labels with our COMNet can achieve higher mIoU than the original CAMs without matching for enhancement. The improved initial response maps can facilitate in generating better pixel-wise pseudo ground truths for training the fully-supervised semantic segmentation model.

\begin{table}[]
	\begin{center}
		\scalebox{1.0}{
			\begin{tabular}{ccc|cc}
				\toprule  
				\toprule  
				Methods & Backbone &  Saliency & $\mathit{val}$ & $\mathit{test}$\\
				\midrule  
				AdvEra \ & VGG16 & \ding{51} & 55.0 & 55.7\\
				MCOF \ & ResNet101 & \ding{51} & 60.3 & 61.2\\
				DSRG \ & ResNet101 & \ding{51} & 61.4 & 63.2\\
				AffinityNet \ & ResNet-38 & - & 61.7 & 63.7\\
				IRNet \ & ResNet50 & - & 63.5 & 64.8\\
				OAA \ & ResNet101 & \ding{51} & 65.2 & 66.4\\
				FickleNet \ & ResNet101 & \ding{51} & 64.9 & 65.3\\
				SEAM \ & ResNet38 & - & 64.5 & 65.7\\
				Mining \ & ResNet101 & - & 66.2 & 66.9\\
				CIAN \ & ResNet101 & - & 64.3 & 65.3\\
				\midrule  
			    \rowcolor{mygray}COMNet (ours) & ResNet50 & - & \textbf{67.1} & \textbf{67.6} \\
				\rowcolor{mygray}COMNet (ours) & ResNet101 & - & \textbf{67.8} &  \textbf{68.1} \\
				\bottomrule 
		\end{tabular}}
	\end{center}\caption{Performance comparisons with the state-of-the-art weakly supervised semantic segmentation methods on the PASCAL VOC 2012 dataset.}\label{table7}
\end{table}

\begin{table}[]
	\begin{center}
		\scalebox{1.0}{
			\begin{tabular}{cc|c}
                \toprule  
                Methods   & Backbone & $\mathit{val}$\\
                \midrule  
                BFBP \  & VGG16 & 20.4\\
                SEC \  & VGG16 & 22.4\\
                DSRG \  & VGG16 & 26.0\\
                IRNet* \  & ResNet38 & 32.6\\
                IAL* \  & VGG16 & 27.7\\
                SEAM* \  & ResNet38 & 31.9\\
                \midrule  
                \rowcolor{mygray}COMNet (ours) & ResNet50 & \textbf{33.9}\\
                \bottomrule 
            \end{tabular}}
	\end{center}\caption{Performance comparisons on COCO dataset. * means our reproduced results since the original papers do not report the results.}\label{table8}
\end{table}

\subsection{Comparison with State-of-the-arts}

Finally, we compare our framework with state-of-the-art methods including: AdvEra~\cite{advera}, MCOF~\cite{MCOF}, DSRG~\cite{DSRG}, AffinityNet~\cite{affinitynet}, IRNet~\cite{irn}, OAA~\cite{OAA}, FickleNet~\cite{ficklenet}, SEAM~\cite{seam}, Mining~\cite{mining} and CIAN~\cite{cian} on the PASCAL VOC 2012 dataset including both the validation set and the testing set.
Tab~\ref{table7} compares our method to previous approaches, our method presents the state-of-the-art performance using only image-level labels with ResNet-50 backbone. When we replace it with ResNet-101 network, we can further boost our performance to the new state-of-the-art. Note that, our performance elevation does not stem from the improved saliency detector. Moreover, we also compare BFBP~\cite{saleh2016built}, SEC~\cite{kolesnikov2016seed}, DSRG~\cite{DSRG}, IAL~\cite{wang2020weakly}, IRNet~\cite{irn} and SEAM~\cite{seam} on MS-COCO dataset. As reported in Tab~\ref{table8}, our model achieves the best
mIoU score on the validation set, which further proves the superiority of our model.
The performance improvement mainly comes from the two matching modules, which produces better CAMs for the segmentation task.

\subsection{Qualitative Results}\label{Qualitative Results}

\noindent \textbf{Failure case.} To help better understand the network interpretability, we show a pair of hard examples for failure case. As is shown in Fig~\ref{fig7}, the objects in different classes are of similar color and texture. The inter-matching module fails to yield the co-occurrent masks and the intra-matching module wrongly matches with other object features. Therefore, the results are unsatisfactory. However, the paired images are randomly sampled in each training epoch throughout the dataset, the abundant guidance ease the pressure in learning this hard example.
Hence, our framework is robust to the quality of weakly supervised semantic segmentation.

\vspace{1ex}

\noindent \textbf{Segmentation predictions.} We visualize some of the predictions on PASCAL VOC 2012 validation set of both the baseline and our COMNet, as is shown in Fig~\ref{fig8}. We can observe that our method can help to expand and discover more comprehensive object regions and make more accurate predictions on objects. Matching between images can help the network enhance the related representations with each other and self-matching can learn more consistent representations.

\section{Conclusion}
In this paper, we propose a \textbf{COMNet} for weakly supervised semantic segmentation from a matching perspective. The novel paired images inter-matching and single images intra-matching modules are proposed to help the network discover more object regions. Experimental results show the effectiveness of our method, which outperforms the baseline model and achieves the new state-of-the-art performance.
This paper pushes the frontier of the weakly supervised semantic segmentation and reduces its performance gap with the supervised segmentation. This work is also a general regularization approach that may benefit to weakly supervised object detection. The ultimate research vision is to potentially relieve the burden of human annotations on training data. This effort may reduce the cost, balance human bias, accelerate the evolution of machine perception technology, and help us to understand how to enable learning with minimal supervision.

\bibliographystyle{IEEEtran}
\bibliography{mybib}

\end{document}